\documentclass{article}
\usepackage{spconf,amsmath,graphicx}
\usepackage{siunitx}

\usepackage{mwe} 

\title{Consistent Recurrent Neural Networks for 3D Neuron Segmentation.}
\name{Felix Gonda, Donglai Wei, Hanspeter Pfister}
\address{Harvard University\\Cambridge, MA}

\begin{document}
%
\maketitle
\begin{abstract}
We present a recurrent network for 3D reconstruction of neurons that sequentially generates binary masks for every object in an image with spatio-temporal consistency.  Our network models consistency in two parts:   (i) local, which allows exploring non-occluding and temporally-adjacent object relationships with bi-directional recurrence. (ii) non-local, which allows exploring long-range object relationships in the temporal domain with skip connections.  Our proposed network is end-to-end trainable from an input image to a sequence of object masks, and, compared to methods relying on object boundaries, its output does not require post-processing. We evaluate our method on three benchmarks for neuron segmentation and achieved state-of-the-art performance on the SNEMI3D~\cite{snemi} challenge. 
\end{abstract}


\section{Introduction}
\label{sec:intro}
The field of connectomics aims to reconstruct the brain's wiring diagram by mapping the neural connections at the level of individual synapses. A reconstruction of neurons' anatomical structures and the synaptic connectivity between them can help neuroscientists better understand the structure and function of the brain~\cite{morgan2013not}. Recent advances in Electron Microscopy (EM) technology make it possible to generate terabytes of brain images at the nanometer scale on an hourly basis \cite{richard2016imaging}.  Thus, efficient and accurate neuron segmentation methods are required to process these images.

Previous approaches~\cite{funke,superhuman} addressed neuron segmentation in multiple steps.  First, a convolutional neural network (CNN) is applied to predict neurons' instance boundaries or affinities. A watershed transform~\cite{zwatershed} is then used to generate an initial 3D over-segmentation, where segments are further merged based on hand-crafted or learned features. However, the CNNs are applied independently on each local sub-volume without any shape information about neighboring regions.  Thus, when image artifacts or unexpected appearance occurs in the input volume, the CNNs make wrong predictions, leading to merging and splitting errors in the final segmentation. 

More recent approaches~\cite{ffn,meirovitch2016multi} treat neuron segmentation as video object tracking along the z-axes of a 3D image volume. These approaches alleviate the appearance problem by learning spatial features through recurrent neural network models. These models segment one object and generate the object mask for the input sub-volume with additional input features from the previous inference step. However, these methods are computationally expensive. For instance, on a 9x9x20nm image volume of the zebra finch songbird dataset, one inference step of the flood-filling networks~\cite{ffn} costs 41.05 EFLOPS with a wall time of 3.15 hrs. 

In terms of spatial consistency, the affinity-based methods~\cite{funke,superhuman} learn them all-in-one, while the tracking-based methods~\cite{ffn,meirovitch2016multi} learn them one-by-one without exploiting the pairwise non-occluding relationships. In terms of temporal consistency during inference, the tracking-based methods run the inference only in forward and backward directions, without long-range consistency. As such, image slices with severe artifacts can lead to broken segmentation. Our approach is inspired by the idea of spatio-temporal recurrence previously demonstrated for video object segmentation~\cite{rvos}, where objects are segmented by propagating masks from a reference frame.  We argue that exploiting the reference frame's semantic meaning has benefits in learning long-range object relationships in sequences.  As such, we aim to extend spatial and temporal recurrence to explore local and non-local relationships between objects in the volume to connect broken segments and form more complete object masks. 

\begin{figure*}[t]
\centering 
\begin{minipage}[b]{1.0\linewidth}
   \centerline{\includegraphics[clip, trim=2.2cm 0.0cm 2.2cm 0.0cm, width=\textwidth]{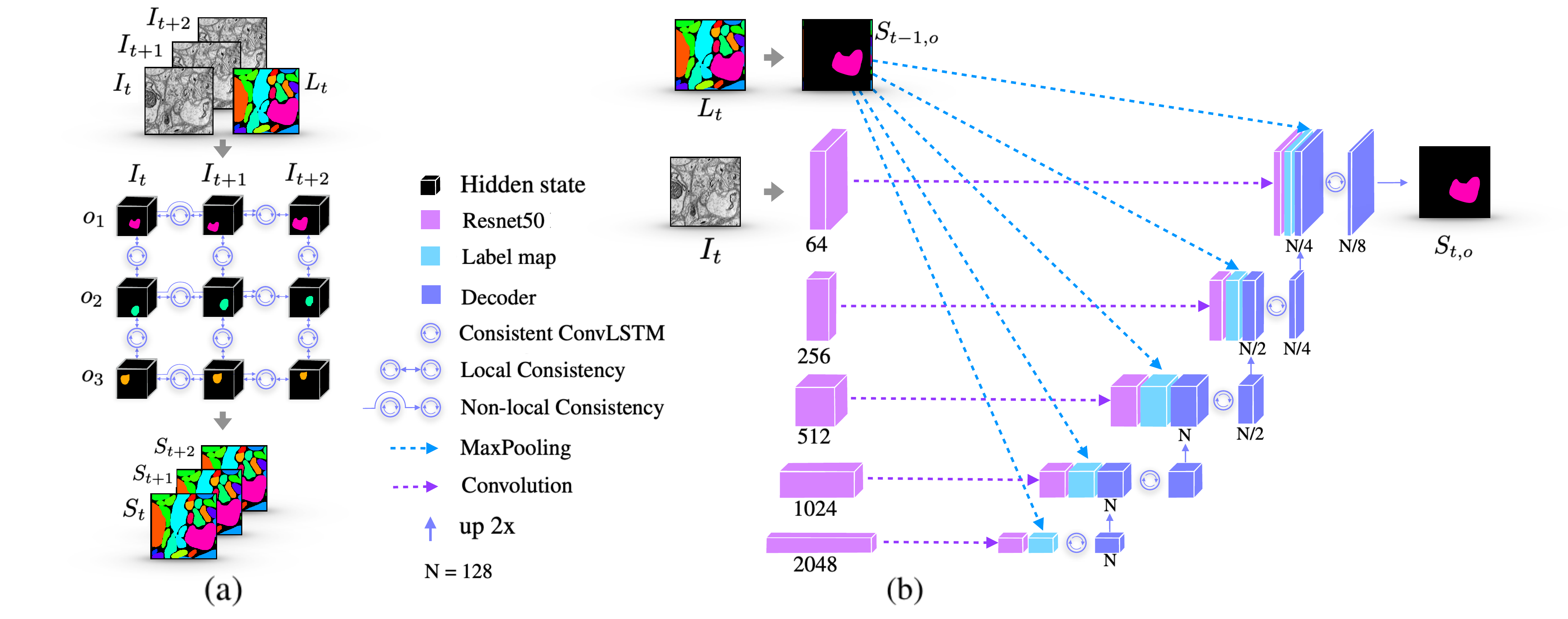}}
\end{minipage}
\caption{
 Our proposed architecture for multi-neuron segmentation.  (a) A network of our Consistent ConvLSTM extract neurons from a sequence $I$ based on a single label map $L_{t}$. (b) A single forward pass of our model predicting the first object $S_{t,o}$ of a frame given the mask of the object from the previous frame $S_{t-1,o}$.  When $t$ is the reference frame,  $S_{t-1,o} = L_{t,o}$
}
\label{fig:architecture}
\end{figure*}

Therefore, we propose a recurrent model that performs a sequential analysis of the input image volume to deal with complex object distributions and generate consistent predictions. We create a new recurrent module based on Convolutional LSTM~\cite{lstm}, as a building block of our network, to model local and non-local object consistency. Given a single label map, our model can segment multiple objects without post-processing. We require no intermediate representation for our model; as such, its training is performed in an end-to-end fashion.  Accuracy results on three connectomics datasets show that our method performs similar to state-of-the-art methods and surpasses human accuracy for 3D neuron segmentation on the SNEMI3D \cite{snemi} dataset. Our results also demonstrate that our method consistently produces object masks that are robust to image artifacts.

\section{Methodology}\label{sec:method}
\subsection{Consistent Recurrent Neural Network}
We propose a recurrent network, depicted in Fig \ref{fig:architecture}, based on an encoder-decoder architecture to solve the task of neuron segmentation.  The network incorporates spatio-temporal recurrence with our Consistent ConvLSTM (CConvLSTM) module defined in Section~\ref{sec:recurrence}.  Our recurrence is configured in the spatial domain (rows) to represent the object instances in a frame and in the temporal domain (columns) to represent frames. The model's input consists of 3D patches transformed into sequences of images along the z-direction of the input volume and a single label map for each sequence. For refinement, each sequence is accompanied by a channel carrying initial object mask estimation. The output of our model is a set of predicted masks $S = \{S_{t,o}, S_{t,o+1},..., S_{t,o+M}\}$, where $S_{t,o}$ is the predicted mask of object $o$ at frame $t$ and $M$ is the number of objects.
\begin{figure}[htb]
\centering 
\begin{minipage}[b]{1.0\linewidth}
   \centerline{\includegraphics[clip, trim=0.5cm 1.5cm 0.5cm 1.0cm, width=\textwidth]{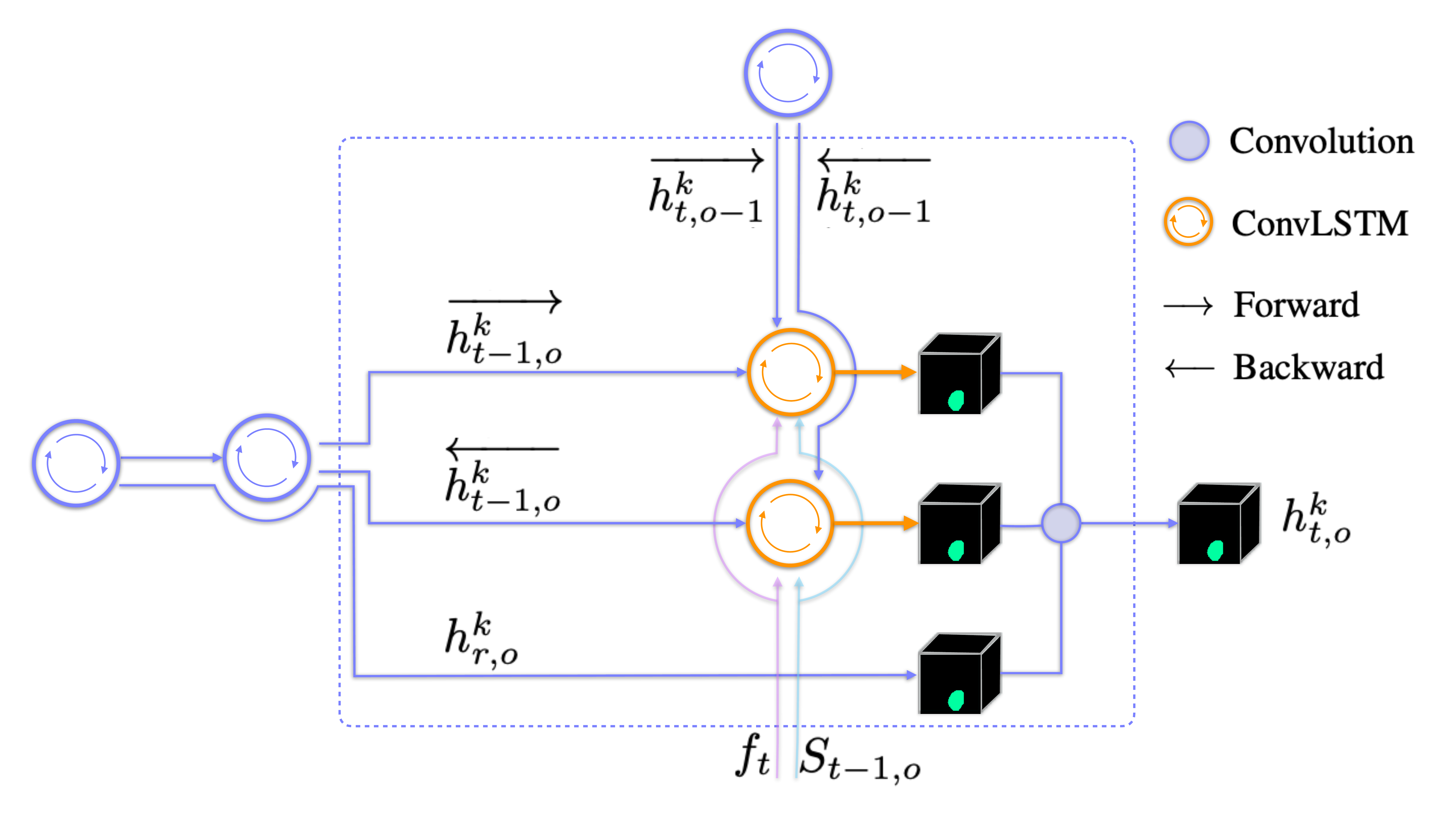}}
\end{minipage}
\caption{
A CConvLSTM module models local object consistency with two ConvLSTM layers to process input in the forward and backward directions.  Non-local object consistency is modeled with a skip connection that integrates the reference hidden state $h^{k}_{r,o}$
}
\label{fig:recurrence}
\end{figure}



\subsection{Spatio-Temporal Consistency Module}\label{sec:recurrence}
To model object consistency, we introduce a new recurrent module, CConvLSTM shown in Fig. \ref{fig:recurrence}, as a building block of our network. This module models local object consistency by combining two ConvLSTM layers that process input in the forward and backward directions. The output of the two layers is convolved with the reference hidden state to model non-local object consistency. The bi-directional flow analysis in the temporal domain has been shown in \cite{deepbilstm} to yield improved predictive performance. 

Therefore, given a sequence and a set of objects propagated from a reference frame $r$, the output $h^{k}_{t,o}$ of the $k^{th}$ CConvLSTM layer for object ${o}$ at frame $t$ is computed as:
\begin{align} \label{eq:recurrence1}
h_{input}& = [ B_{2}(h^{k-1}_{t,o})|f^{k'}_{t}|S_{t-1,o}]\\
\overrightarrow{h_{state}} & = [ \overrightarrow{h^{k}_{t,o-1}} |  \overrightarrow{h^{k}_{t-1,o}} ]\\
\overleftarrow{h_{state}} & = [ \overleftarrow{h^{k}_{t,o-1}} |  \overleftarrow{h^{k}_{t-1,o}} ]\\
h^{k}_{t,o}& = CConvLSTM( h_{input},\overrightarrow{h_{state}},\overleftarrow{h_{state}},h^{k}_{r,o})\label{eqn:hidden_state}
\end{align}
where,  B2 is the bilinear upsampling of the output of the previous CConvLSTM layer by a factor of $2$. $f^{k}_{t}$ is the features from the encoder at frame $t$ and $f^{k'}_{t,k}$ is the projection of $f^{k}_{t}$ to lower dimension via a convolution layer.  $S_{t-1,o}$ is the predicted segmentation mask of the object from the previous frame. $\overrightarrow{h^{k}_{t,o-1}}$ and $\overleftarrow{h^{k}_{t,o-1}}$ are the forward and backward components of the spatial hidden state representation of the previous object. $\overrightarrow{h^{k}_{t-1,o}}$ and $\overleftarrow{h^{k}_{t-1,o}}$ are the forward and backward components of the temporal hidden state of the object from the previous frame. For the first hidden state $h^{0}_{t,o}$ of the object, we use the segmentation mask from the reference $S_{r,o}$ frame and the zero matrix for the spatial hidden state. We assess the importance of the local and non-local object consistency in Section \ref{sec:results}.


\subsection{Encoding Path}
The encoder, illustrated in pink in Fig.~\ref{fig:architecture}, is a Resnet50 \cite{resnet} that is truncated at the last convolution layer.  The encoder learns to extract features, $f = \{f_{t}, f_{t+1}, f_{t+2}, f_{t+3}, f_{t+4}\}$, from an RGB image ${x \in R^{h \times w \times 3}}$ corresponding to the output of Resnet blocks. $f_{t}$ corresponds to the output of the deepest block, and $f_{t+4}$ corresponds to the output of the block whose input is the image. A convolution operation is used to extract the features from each block.

\subsection{Decoding Path}
The decoder, shown in dark blue in Fig.~\ref{fig:architecture} (b), is a hierarchical recurrent architecture of CConvLSTMs leveraging the different resolutions of the input features $f$ and label map $L_{t}$. The label map is extracted with a series of down-sampling operations corresponding to the resolution of features $f$ as shown in blue in Fig.~\ref{fig:architecture} (b). The output of each CConvLSTM is subsequently merged with corresponding encoder features and object masks, which allows the decoder to reuse low-level features and refine the final segmentation. The decoder applies equation \ref{eqn:hidden_state} in chains for the number of CConvLSTMs.  The decoder's output is a set of $ M $ predictions per image. $M$ is the number of objects propagated and is always constant per sequence to ensure an object mask will be empty if it disappears in subsequent frames.  The constant $M$ also ensures the  predicted mask in the temporal domain is consistent with the spatial recurrence. 

\subsection{Implementation Details}
Since the number of propagated objects is always equal to the number of predicted objects, we estimate the parameters of our model by optimizing an objective function based on the Hungarian algorithm \cite{hungarian} using soft intersection over union. Thus, given a sequence of $m$ predicted and $g$ ground truth masks of length $N$, the loss can be expressed as:
$$
sIoU(m,g)=1.0 - \frac{\sum_{i=1}^{N}m_{i}g_{i}}{(\sum_{i=1}^{N}m_{i} + g_{i}  - m_{i}g_{i} )}\\
$$

The network is trained using the Adam optimizer with a learning rate of $10^{-6}$ and a batch size of 1 over 40 epochs. For the early ten epochs, the ground-truth mask of objects from the previous frame is included as an additional input channel to our CConvLSTM. For the remaining 30 epochs, the object's inferred mask is used to fine-tune the model, thus allowing the model to learn to fix errors that may occur at inference time.  The training was carried out on a single NVIDIA Titan X GPU with 12GB RAM for 24 hours. 

\section{Experiments}\label{sec:experiments}
\subsection{Setup}
We evaluate our method's efficacy with experiments on three EM datasets from different species, as described in Table \ref{tab:datasets}. The ($x, y, z$) dimensions of each datasets in voxels is: ($1024\times 1024\times 100$), ($500\times 500\times 500$), and ($1024\times1024\times105$) for SNEMI~\cite{snemi}, FIBSEM~\cite{fibsem}, and FIBER respectively. The FIBER dataset is in-house generated, the FIBSEM is public, and the SNEMI is a benchmark for the SNEMI challenge.  

\begin{table}[ht]
\centering
\resizebox{\columnwidth}{!}{\begin{tabular}{@{}ccc}
  \hline
        \textbf{Name} & \textbf{Species (region)} & \textbf{Volume}  \\ \hline
        SNEMI~\cite{kasthuri2015saturated} & Mouse (Cortex) & \SI[product-units=single]{6x6x3}{\micro\meter^3}  \\
         FIBSEM~\cite{fibsem} & Fruit Fly (Medulla) & \SI[product-units=single]{5x5x5}{\micro\meter^3}  \\ 
        FIBER (in-house) & Mouse (Cerebellum) & \SI[product-units=single]{8x8x3}{\micro\meter^3}  \\\hline
\end{tabular}}
\caption{The list of datasets used to evaluate our system. Each consists of a training and testing volume. The training volume is split 80\% for training and 20\% for validation. 
}
\label{tab:datasets}
\end{table}

Our baseline model implements spatio-temporal (ST) recurrence. We consider three options to analyze the importance of consistency: (i) local consistency model STL (non-local consistency not used), (ii) non-local consistency model STN (local consistency not used), and (iii) a combined model STC (both local and non-local consistency used). We compare our best model against the SNEMI challenge leaderboard and the affinity-based model, waterz~\cite{funke}. We use the author's publicly available waterz method implementation and use the original paper's suggested hyper-parameters.

During inference, we process the input 3D volume patches in an overlapping manner, starting with a labeled frame.  The segmentation of the first frame is initialized with its corresponding ground-truth. A watershed transform output is used to initialize the first frame's segmentation for volumes where no ground-truth exists. 


We analyze our results using the Adaptive Rand Index (ARI)~\cite{randindex} used by the SNEMI challenge~\cite{snemi} to be consistent. The ARI measures the similarity between two data clusters.  The error is defined as one minus the maximal F-score of the Rand index. A lower ARI score corresponds to better segmentation quality. 

\subsection{Results}\label{sec:results}
\begin{figure}[t]
\centering 
\begin{minipage}[b]{1.0\linewidth}
   \centerline{\includegraphics[clip, trim=10.0cm 0.0cm 10.0cm 0.0cm, width=\textwidth]{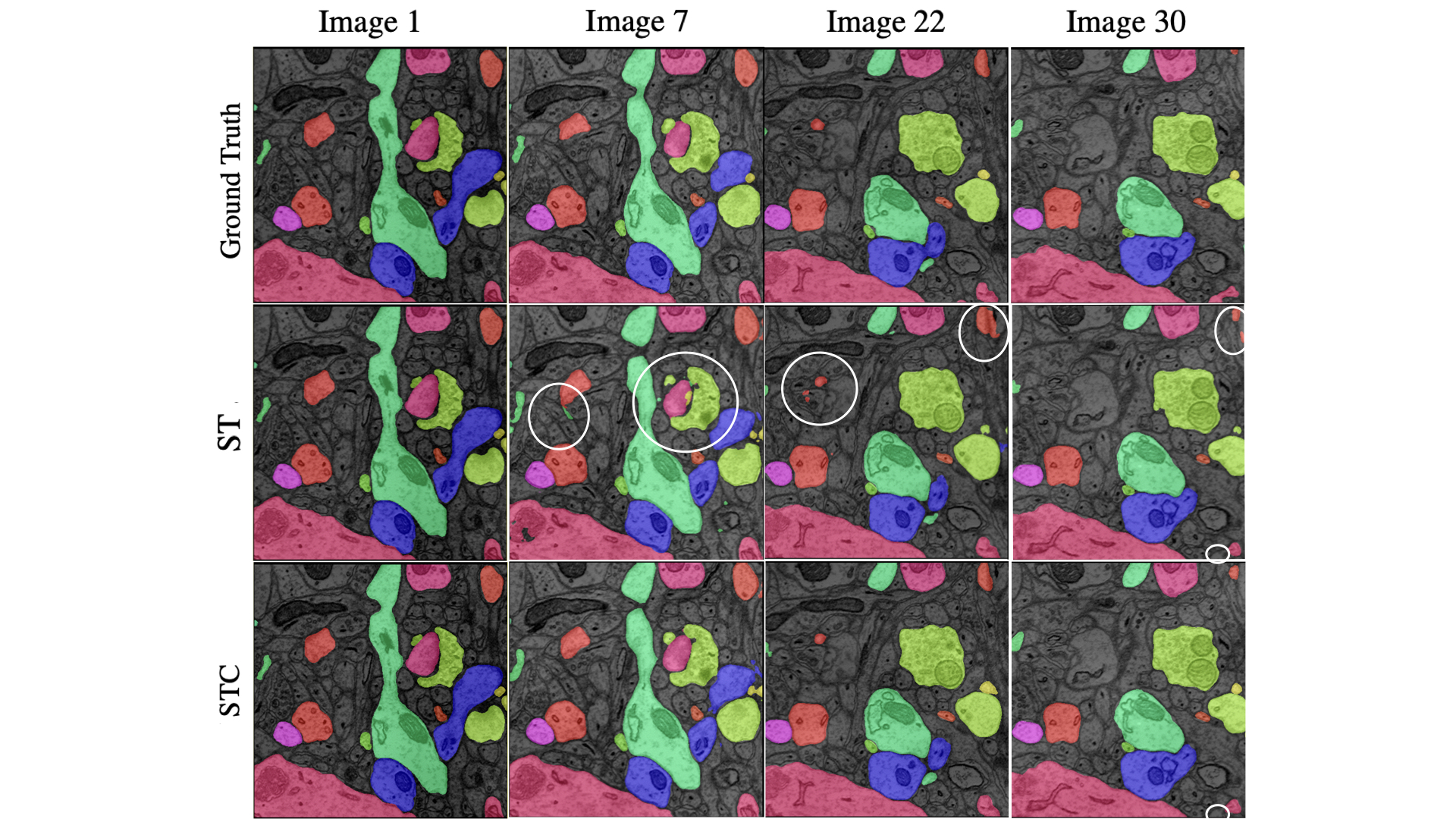}}
\end{minipage}
\caption{
We extract 15 neurons from a sequence of 30 images from the SNEMI dataset using the ST and our STC models.  In comparison to the ground truth, errors are circled in white.
}
\label{fig:ablation}
\end{figure}

In Fig.~\ref{fig:ablation}, we show the segmentation of 15 objects by the ST and STC models in a validation sequence compared to ground-truth. As shown by the white circles, propagation errors occur when consistency is not used. These observations are further confirmed with the quantitative evaluations, shown in Table \ref{tab:ablation_studies}. The non-local consistency model (STN) is faster and connects distant segments better, and when combined with local consistency, produces the lowest ARI value as demonstrated by the STC model shown in Table~\ref{tab:ablation_studies}.

\begin{figure}[t]
\centering 
\begin{minipage}[b]{1.0\linewidth}
   \centerline{\includegraphics[clip, trim=10.0cm 0.0cm 10.0cm 0.0cm, width=\textwidth]{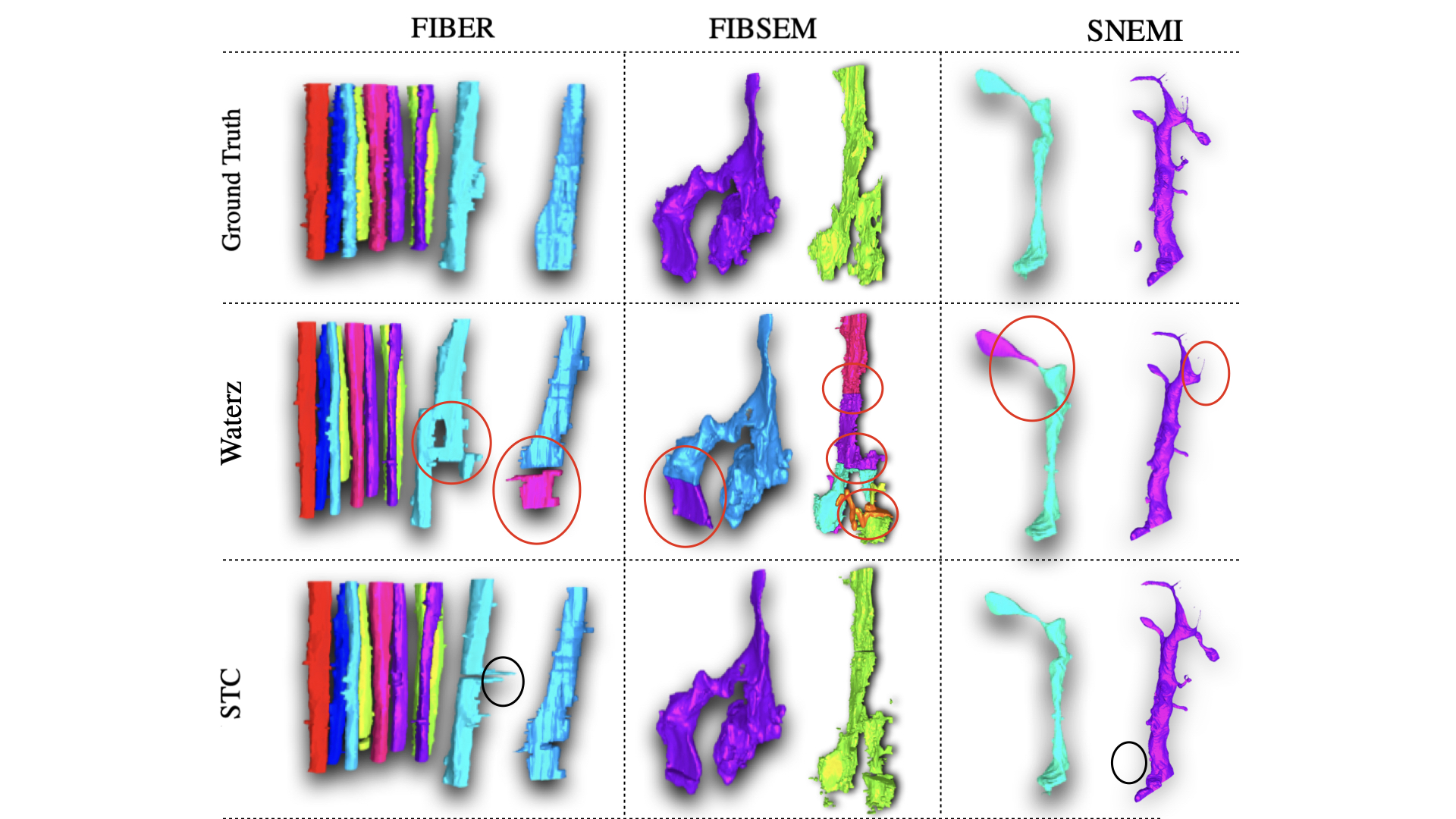}}
\end{minipage}
\caption{
Qualitative comparison of our STC to the waterz~\cite{funke} method. Errors shown in red and black circles.
}
\label{fig:qualitative}
\end{figure}


\begin{table}[h]
\centering
\resizebox{\columnwidth}{!}{\begin{tabular}{@{}lcc}
  \hline
 Model & Accuracy (ARI) & Inference Time (seconds)  ~~ \\
\hline
ST \cite{rvos}      & 0.13  & 520.0 \\
STL & 0.082 & 640.0 \\
STN & 0.045 & 605.0 \\
STC & 0.035 & 660.0 \\
  \hline
\end{tabular}}
\caption{An assessment of local (L) and non-local (N) consistency compared to baseline ST \cite{rvos} model in terms of ARI (lower is better) and inference time in seconds. 
}
\label{tab:ablation_studies}
\end{table}


\begin{table}[th]
\centering
\resizebox{\columnwidth}{!}{\begin{tabular}{@{}lccc}
  \hline
 Method/Dataset & ~SNEMI3D~ & ~FIBSEM~& ~FIBER~ \\
\hline
PNI~\cite{superhuman}  &  \textbf{0.024} & N/A& N/A \\
FFN~\cite{ffn}  &  \textbf{0.029} & N/A& N/A \\
\textbf{STC (ours)} & \textbf{0.035} & \textbf{0.106} & \textbf{0.091}\\ 
(human values)  & 0.059 & N/A & N/A\\ 
waterz~\cite{funke} & 0.072 & 0.163 & 0.210\\
  \hline
\end{tabular}}
\caption{Our STC achieved third place ranking on SNEMI3D~\cite{snemi} leaderboard and outperform waterz~\cite{funke} on all three datasets. 
}
\label{tab:benchmark}
\end{table}


For a fair comparison against state-of-the-art methods, we start from the same watershed output as the waterz\cite{funke} method. In Fig.~\ref{fig:qualitative}, we highlight in red circles errors in the waterz method that are addressed by our spatio-temporal consistency. Although our method is robust to segmentation breakage, some artifacts may not be fully corrected as shown in black if the watershed is bad.  In this case, fine-tuning the watershed before applying our model is an alternate solution. In table~\ref{tab:benchmark}, we compare our best model STC against the SNEMI3D \cite{snemi} challenge leaderboard. Our method achieved 0.035 ARI score, surpassing the human accuracy, and is currently ranked third after PNI~\cite{superhuman} and FNN~\cite{ffn}. Our method also consistently outperformed the waterz\cite{funke} method on all three datasets. We attribute our method's strength to its ability to learn long-range object relationships, critical for connectomics data. This is demonstrated, in Figure~\ref{fig:qualitative}, with the presence of segment splits in waterz and their absense in our method. In terms of inference time on the SNEMI dataset, our method requires $660$ seconds compared to 61 minutes for the FFN~\cite{ffn} model. 

\section{Conclusion}\label{sec:conclusion}
The proposed recurrent model learns to deal with complex object distribution across long sequences and produces segments that are consistent with each other. By training our model end-to-end, we eliminated intermediate representations that could potential introduce errors such as in the affinity-based methods.  Our model is also suited for interactive segmentation where object propagation is driven by the user. Therefore, we plan to incorporate our model in the proofreading of neurons to help correct automatic segmentation errors.

\section{Compliance with Ethical Standards}
\label{sec:ethics}
This is a numerical simulation study for which no ethical approval was required.

\section{Acknowledgments}
\label{sec:acknowledgments}
This work was partially supported by NSF grant IIS-1607800.

\bibliographystyle{IEEEbib}
\bibliography{root}

\end{document}